\documentclass[letterpaper]{article}
\usepackage{ijcai13}
\usepackage{named}
\usepackage{times}

\usepackage{amsmath}
\usepackage{listings}
\usepackage{wrapfig}

\long\def\BOC#1\EOC{\message{(Commented text )}}
\long\def\BOCC#1\EOCC{\message{(Commented text )}}
\long\def\BOCCC#1\EOCCC{\message{(Commented text )}}
\long\def\optional#1{\empty}
\long\def\NB#1{}

\def\ar{\leftarrow}
\def\beq{\begin{equation}}
\def\eeq#1{\label{#1}\end{equation}}
\def\ba{\begin{array}}
\def\ea{\end{array}}
\def\bi{\begin{itemize}}
\def\ei{\end{itemize}}
\def\i#1{\hbox{\it #1\/}}
\def\mi#1{\mathit{#1}}
\def\mu#1{\mathit{\underline{#1}}}

\def\sm{\hbox{\rm SM}}
\def\fsm{\hbox{\rm SM}}
\def\no{\i{not}}
\def\rar{\rightarrow}
\def\lrar{\leftrightarrow}

\def\v{\widehat}
\def\bC{{\bf{c}}}
\def\vbC{{\v{\bf{c}}}}

\def\dia{\diamond} 

\def\false{\hbox{\sc false}}
\def\true{\hbox{\sc true}}

\def\mvis{\!=\!}

\newtheorem{thm}{Theorem}

\newtheorem{example}{Example}

\title{Functional Stable Model Semantics and \\ Answer Set
  Programming Modulo Theories}

\author{Michael Bartholomew and Joohyung Lee\\ 
School of Computing, Informatics and Decision Systems Engineering \\
Arizona State University, Tempe, USA \\
{\tt \{mjbartho, joolee\}@asu.edu}
}

\begin{document}

\maketitle

\begin{abstract}
Recently there has been an increasing interest in incorporating
``intensional'' functions in answer set programming. Intensional
functions are those whose values can be described by other functions
and predicates, rather than being pre-defined as in the standard
answer set programming. We demonstrate that the functional stable model
semantics plays an important role in the framework of ``Answer Set
Programming Modulo Theories (ASPMT)'' ---a tight integration of answer
set programming and satisfiability modulo theories, under which
existing integration approaches can be viewed as special cases where
the role of functions is limited. We show that ``tight'' ASPMT
programs can be translated into SMT instances, which is similar to the
known relationship between ASP and SAT.
\end{abstract}

\section{Introduction} \label{sec:intro}

In answer set programming (ASP), variables are understood in terms of
grounding, and this limits answer sets to Herbrand models, under which
interpretations of functions are pre-defined:
every ground term represents itself, and is distinct from each
other. Nonmonotonicity of the stable model semantics is related to
minimizing the extents of predicates, but has nothing to do with
functions, which forces us to represent the concept of nonBoolean
fluents indirectly in terms of predicates, and not by functions. 
A drawback of such representation is that grounding often becomes a
bottleneck in computation as the value domain of a nonBoolean fluent
gets large. 

There are two recent groups of work to extend the stable model
semantics, each of which focuses only on one aspect of the issues
above.
One group is interested in enriching the modeling language by
incorporating  ``intensional''
functions~\cite{cabalar11functional,lifschitz12logic,bartholomew12stable,balduccini12aconservative}.
Intensional functions are functions whose values can be described
by other functions and predicates, rather than being pre-defined. Such
semantics allow us to represent nonBoolean fluents by intensional
functions without having to rely on predicates. 

The other group of work focuses on improving the computational
efficiency by
integrating ASP with other declarative computing paradigms, such as
constraint processing, satisfiability modulo
theories, or mixed integer
programming~\cite{gebser09constraint,balduccini09representing,janhunen11tight,liu12answer},
and exploiting the efficient constraint processing techniques on
functions (or called ``variables'' in CSP/SMT) without having to
generate a large set of ground instances. 
Constraint variables are functions that are mapped to
values in their domains. In SMT with difference logic or linear
constraints, arithmetic variables are functions that are mapped to
numbers.
However, these functions are not as expressive as
intensional functions. 

This paper combines the advantages of the two groups of
work resulting in the framework of {\em Answer Set Programming Modulo
Theories (ASPMT)}---a tight integration of answer set programming and
satisfiability modulo theories, where functions are as expressive as
predicates in answer set programming, and can be computed efficiently
without having to ground w.r.t. their value domains. 
The existing languages in the second group can be viewed as special
cases of this language.  A fragment of ASPMT can be translated
into the language of SMT, allowing SMT solvers to be used for
computation.

Section~\ref{sec:smif} reviews the functional stable model semantics by
\citeauthor{bartholomew12stable}~[\citeyear{bartholomew12stable}]. 
Section~\ref{sec:reduct} presents a reformulation of this semantics
in terms of grounding and reduct for infinitary ground formulas.
In Section~\ref{sec:aspmt} we define the concept of ASPMT as a special
case of the functional stable model semantics, and
presents a syntactic class of ASPMT instances that can be translated
into SMT instances, which in turn allows us to use SMT solvers to
compute ASPMT. In Section~\ref{sec:comparison} we compare
ASPMT with other similar approaches, clingcon programs and ASP(LC)
programs.

\section{Functional Stable Model Semantics}\label{sec:smif}


\subsection{Review of the Bartholomew-Lee Semantics}

We review the stable model semantics of intensional functions
from~\cite{bartholomew12stable}.
Formulas are built the same as in first-order logic. A signature
consists of {\em function constants} and {\em predicate constants}. 
Function constants of arity $0$ are called {\em object constants}, 
and predicate constants of arity $0$ are called {\em propositional
  constants}. 

Similar to circumscription, for predicate symbols (constants or
variables) $u$ and $c$, expression $u\le c$ is defined as shorthand for
\hbox{$\forall {\bf x}(u({\bf x})\rar c({\bf x}))$}.
Expression $u=c$ is defined as
\hbox{$\forall {\bf x}(u({\bf x})\lrar c({\bf x}))$}
if $u$ and $c$ are predicate symbols, and
$\forall {\bf x}(u({\bf x})=c({\bf x}))$
if they are function symbols.
For lists of symbols ${\bf u}=(u_1,\dots,u_n)$ and
${\bf c}=(c_1,\dots,c_n)$, expression ${\bf u}\le {\bf c}$ is defined
as $(u_1\le c_1)\land\dots\land (u_n\le c_n)$, 
and similarly, expression ${\bf u} = {\bf c}$ is defined
as $(u_1 = c_1)\land\dots\land (u_n = c_n)$.
Let $\bC$ be a list of distinct predicate and function constants, and
let $\vbC$ be a list of distinct predicate and function variables
corresponding to~$\bC$. 
By $\bC^\mi{pred}$ ($\bC^\mi{func}$, respectively) we mean the
list of all predicate constants (function constants, respectively) in
$\bC$, and by $\vbC^\mi{pred}$ ($\vbC^\mi{func}$, respectively) the
list of the corresponding predicate variables (function
  variables, respectively) in $\vbC$.

For any formula $F$, expression $\fsm[F;\ \bC]$ is defined as
\[
   F\land\neg\exists \vbC(\vbC<\bC\land F^*(\vbC)),
\]
where $\vbC<\bC$ is shorthand for 
$(\vbC^\mi{pred}\le \bC^\mi{pred})\land\neg (\vbC=\bC)$, 
and $F^*(\vbC)$ is defined recursively as follows.
\begin{itemize}
\item When $F$ is an atomic formula, $F^*$ is $F'\land F$ where $F'$
  is obtained from $F$ by replacing all intensional (function and
  predicate) constants ${\bf c}$ in it with the corresponding
  (function and predicate) variables from $\v{\bC}$;


\item  $(G\land H)^* = G^*\land H^*$;\ \ \ \ \
       $(G\lor H)^* = G^*\lor H^*$;

\item  $(G\rar H)^* = (G^*\rar H^*)\land (G\rar H)$;

\item  $(\forall x G)^* = \forall x G^*$;\ \ \ \ \
       $(\exists xF)^* = \exists x F^*$.
\end{itemize}
(We understand $\neg F$ as shorthand for $F\rar\bot$; $\top$ as
$\neg \bot$; and $F\lrar G$ as $(F\rar G)\land (G\rar F)$.)
Members of $\bC$ are called {\sl intensional} constants.

When $F$ is a sentence, the models of $\fsm[F; {\bf c}]$ are called
the {\em stable} models of~$F$ {\em  relative to} ${\bf c}$. They are
the models of~$F$ that are ``stable'' on ${\bf c}$.
The definition can be easily extended to formulas of many-sorted
signatures.

This definition of a stable model is a proper generalization of the
one from~\cite{ferraris11stable}, which views logic programs as a
special case of first-order formulas. 

We will often write $G\ar F$, in a rule form as in logic programs, 
to denote the universal closure of $F\rar G$. 
A finite set of formulas is identified with the conjunction of the
formulas in the set.

\begin{example}\label{ex:1}
The following set $F$ of formulas describes the capacity of a container of
water that has a leak but that can be refilled to the maximum amount,
say 10, with the action $\i{FillUp}$.
\[
\ba {rcl}
  \{\i{Amount}_1\mvis x\} &\!\!\ar\!\!& 
      \i{Amount}_0\mvis x\!+\!1 \\
  \i{Amount}_1\mvis 10 &\!\!\ar\!\!& \i{FillUp}\ .
\ea
\]
Here $\i{Amount}_1$ is an intensional function constant, and $x$ is a
variable ranging over nonnegative integers. According to the
semantics from~\cite{bartholomew12stable}, the first rule is a
default rule (or choice rule) standing for 
\beq 
   (\i{Amount}_1\mvis x)\lor\neg(\i{Amount}_1\mvis x)\ \ar\
   \i{Amount}_0\mvis x\!+\!1,
\eeq{default}
and expresses that the amount at next time decreases by
default. However, if $\i{FillUp}$ action is executed (if we add
$\i{FillUp}$ as a fact), this behavior is overridden, and the amount
is set to the maximum value. 

Consider an interpretation $I$ that has the set of nonnegative
integers as the universe, interprets integers, arithmetic functions
and comparison operators in the standard way, and has 
$\i{FillUp}^I = \false$, 
$\i{Amount}_0^I= 6$, 
\hbox{$\i{Amount}_1^I = 5$}. 
One can check that $I$ is a model of $\sm[F; \i{Amount}_1]$.
Consider another interpretation $I_1$ that agrees with~$I$ except
that \hbox{$\i{Amount}_1^{I_1} = 8$}. This is a model of $F$ but not of 
$\sm[F; \i{Amount}_1]$.
Another interpretation $I_2$ that agrees with~$I$ except that 
$\i{FillUp}^{I_2}=\true$, $\i{Amount}_1^{I_2}=10$ 
satisfies $\sm[F;\i{Amount}_1]$.
\end{example}

This example demonstrates the ability to assign a default value to an 
intensional function, which is different from the previous value of
the function.

\section{FSM in terms of Grounding and Reduct} \label{sec:reduct}

Instead of relying on a transformation into second-order logic, 
the definition of a stable model in the previous section can be
characterized in terms of grounding and reduct. 
While the definition in terms of second-order logic is succinct, the
reduct-based definition is more familiar, and tells us some other
insights.

\subsection{Review: Infinitary ground Formulas} 

Since the universe can be infinite, grounding a quantified sentence
introduces infinite conjunctions and disjunctions over the elements 
in the universe. Here we rely on the concept of grounding {\sl
  relative to an interpretation} from
\cite{truszczynski12connecting}. 
The following is the definition of an {\em infinitary ground formula},
which is adapted from~\cite{truszczynski12connecting}. The main
difference between them is that we do not replace ground terms with
their corresponding object names, leaving intensional functions in
grounding. This is essential for defining a reduct for the functional
stable model semantics. 

For each element $\xi$ of the universe $|I|$ of $I$, we introduce a
new symbol $\xi^\dia$, called an {\sl object name}. By $\sigma^I$ we
denote the signature obtained from~$\sigma$ by adding all object names
$\xi^\dia$ as additional object constants. We will identify an
interpretation~$I$ of signature $\sigma$ with its extension to
$\sigma^I$ defined by $I(\xi^\dia)=\xi$.\footnote{%
For details, see \cite{lif08b}.}

We assume the primary connectives to be $\bot$, $\{\}^\land$,
$\{\}^\lor$, and $\rar$. 
Propositional connectives $\land,\lor,\neg,\top$ are considered as
shorthands: $F\land G$ as $\{F,G\}^\land$; $F\lor G$ as
$\{F,G\}^\lor$.

Let $A$ be the set of all ground atomic formulas of
signature~$\sigma$. The sets ${\cal F}_0, {\cal F}_1, \dots$ are
defined recursively as follows:
\begin{itemize}
\item ${\cal F}_0=A\cup\{\bot\}$;
\item ${\cal F}_{i+1} (i\ge 0)$ consists of expressions
  ${\cal H}^\land$ and ${\cal H}^\lor$, for all subset
  ${\cal H}$ of ${\cal F}_0\cup\ldots\cup{\cal F}_i$, and of
  the expressions $F\rar G$, where
  $F,G\in{\cal F}_0\cup\dots\cup{\cal F}_i$.
\end{itemize}

We define
${\cal L}_A^{inf}=\bigcup_{i=0}^{\infty}{\cal F}_i$. We call
elements of ${\cal L}_A^{inf}$ {\em infinitary ground formulas}
of~$\sigma$.  

For any interpretation $I$ of $\sigma$ and any infinitary ground
formula $F$, the definition of satisfaction, $I\models F$, is as
follows:
\begin{itemize}
\item For atomic formulas, the definition is the same as in the standard
  first-order logic;


\item $I\models{\cal H}^\land$ if for every formula $G\in {\cal H}$, 
   $I \models G$;

\item $I\models{\cal H}^\lor$ if there is a formula $G\in {\cal H}$ 
   such that $I\models G$;

\item $I \models G\rar H$ if $I\not\models G$ or $I\models H$.
\end{itemize}


Let $F$ be any first-order sentence of a signature $\sigma$, and let
$I$ be an interpretation of~$\sigma$. By $gr_I[F]$ we denote the
{\em infinitary ground formula w.r.t.~$I$}, which is obtained from $F$
by the following process:
\begin{itemize}
\item  If $F$ is an atomic formula, $gr_I[F]$ is $F$; 
\item  $gr_I[F\odot G]= gr_I[F]\odot gr_I[G]\ \ \ \
 (\odot\in\{\land,\lor,\rar\})$;

\item  $gr_I[\forall x F(x)] = 
      \{gr_I[F(\xi^\diamond)] \mid \xi\in |I|\}^\land$;

\item  $gr_I[\exists x F(x)] = 
     \{gr_I[F(\xi^\diamond)] \mid \xi\in |I|\}^\lor$.
\end{itemize}
Note that $gr_I[F]$ is an infinitary ground formula of $\sigma^I$.

\medskip\noindent{\bf Example~\ref{ex:1} continued}\ \ 
{\sl 
Consider again $F$ in Example~\ref{ex:1}, and the same interpretation
$I$ there. $gr_I(F)$ is the following set of formulas: 
\beq
\ba {rcl}
  \{\i{Amount}_1\mvis 0\} &\!\ar\!&
      \i{Amount}_0\mvis 0\!+\!1 \\
  \{\i{Amount}_1\mvis 1\} &\!\ar\!&
      \i{Amount}_0\mvis 1\!+\!1 \\
  \dots \\
  \i{Amount}_1\mvis 10 &\!\ar\!& \i{FillUp}\ .
\ea
\eeq{ex-ground}
}

\subsection{Reduct-Based Definition of SM} 

For any two interpretations $I$, $J$ of the same signature and any
list $\bC$ of distinct predicate and function constants, we write 
$J<^\bC I$ if 
\begin{itemize}
\item  $J$ and $I$ have the same universe and agree on all constants
  not in $\bC$;
\item  $p^J\subseteq p^I$ for all predicate constants $p$ in $\bC$; and
\item  $J$ and $I$ do not agree on $\bC$. 
\end{itemize}

%

The {\em reduct} $F^\mu{I}$ of an infinitary ground formula $F$
relative to an interpretation $I$ is defined as follows: 
\begin{itemize}
\item For each atomic formula $F$, $F^\mu{I}=\bot$ if $I\not\models F$ and
  $F^\mu{I}=F$ otherwise; 

\item $({\cal H}^\land)^\mu{I}=\bot$ if $I\not\models {\cal H}^\land$;
  otherwise \hbox{$({\cal H}^\land)^\mu{I}=\{G^\mu{I} \mid G\in{\cal H}\}^\land$}; 

\item $({\cal H}^\lor)^\mu{I}=\bot$ if $I\not\models {\cal H}^\lor$;
  otherwise \hbox{$({\cal H}^\lor)^\mu{I}=\{G^\mu{I} \mid G\in{\cal H}\}^\lor$}; 

\item $(G\rar H)^\mu{I}=\bot$ if $I\not\models G\rar H$; 
  otherwise 
  \hbox{$(G\rar H)^\mu{I} = G^\mu{I}\rar H^\mu{I}$}. 
\end{itemize}

The following theorem states the reformulation of FSM in terms of
grounding and reduct. 

\begin{thm}\label{thm:reduct-fsm}\optional{thm:reduct-fsm}
Let $F$ be a first-order sentence and $\bC$ a 
list of intensional constants. For any interpretation
$I$ of~$\sigma$, $I\models\sm[F; \bC]$ iff 
\begin{itemize}
\item  $I$ satisfies $F$, and 
\item  every interpretation $J$ such that $J<^\bC I$ 
  does not satisfy $(gr_I[F])^\mu{I}$. 
\end{itemize}  
\end{thm}

\medskip\noindent{\bf Example~\ref{ex:1} continued}\ \ 
{\sl 
Among the ground formulas in~\eqref{ex-ground}, only the implication
\[
 \{\i{Amount}_1\mvis 5\}\ \ar\ \i{Amount}_0\mvis 5\!+\!1\, 
\] 
which stands for 
\[
 (\i{Amount}_1\mvis 5)\lor\neg(\i{Amount}_1\mvis 5)\ \ar\ \i{Amount}_0\mvis 5\!+\!1\, 
\] 
has its antecedent satisfied by $I$, so  
the reduct $(gr_I[F])^\mu{I}$ is equivalent to 
\[
 (\i{Amount}_1\mvis 5)\lor\bot\ \ar\ \i{Amount}_0\mvis 5\!+\!1\, 
\]
No interpretation that is different from $I$ only on $\i{Amount}_1$
satisfies the reduct.
On the other hand, the reduct $(gr_{I_1}[F])^\mu{I_1}$ is equivalent to
\[
 \bot\lor\neg\bot\ \ar\ \i{Amount}_0\mvis 5\!+\!1, 
\]
and other interpretations that are different from $I_1$ only on
$\i{Amount}_1$ satisfy the reduct.
$(gr_{I_2}[F])^\mu{I_2}$ is equivalent to 
\[
\ba l
 \bot\lor\neg\bot\ \ar\ \i{Amount}_0\mvis 5\!+\!1,  \\
 \i{Amount}_1=10\ \ar\ \i{FillUp}
\ea
\]
and $I_2$ is the only interpretation that satisfies the reduct. 

In accordance with Theorem~\ref{thm:reduct-fsm}, $I$ and $I_2$ are the
stable models of $F$ relative to $\i{Amount}_1$, but $I_1$ is not.
}\smallskip

\section{Answer Set Programming Modulo Theories}  
\label{sec:aspmt}

\subsection{ASPMT as a Special Case of FSM} 

Formally, an SMT instance is a formula in many-sorted first-order
logic, where some designated function and predicate constants are
constrained by some fixed background interpretation. SMT is the
problem of determining whether such a formula has a model that expands
the background interpretation~\cite{barrett09satisfiability}.

The syntax of ASPMT is the same as that of SMT. Let $\sigma^{bg}$ be
the (many-sorted) signature of the background theory~$bg$. An
interpretation of $\sigma^{bg}$ is called a {\em background
  interpretation} if it satisfies the background theory. For instance,
in the theory of reals, we assume that $\sigma^{bg}$ contains the set
$\mathcal{R}$ of symbols for all real numbers, the set of arithmetic
functions over real numbers, and the set $\{<, >, \le, \ge\}$ of
binary predicates over real numbers. Background interpretations
interpret these symbols in the standard way.

Let $\sigma$ be a signature that is disjoint from $\sigma^{bg}$.
We say that an interpretation $I$ of $\sigma$ satisfies $F$
w.r.t. the background theory $bg$, denoted by $I\models_{bg} F$,
if there is a background interpretation $J$ of $\sigma^{bg}$ that has
the same universe as $I$, and $I\cup J$ satisfies $F$.
For any ASPMT sentence $F$ with background theory
$\sigma^{bg}$, interpretation $I$ is a stable model of $F$ relative
to~${\bf c}$ (w.r.t. background theory $\sigma^{bg}$) if
$I\models_{bg} \sm[F; \bC]$.



\medskip\noindent{\bf Example~\ref{ex:1} continued}\ \ 
{\sl 
Formula $F$ can be understood as an ASPMT formula with the theory of
integers as the background theory. Arithmetic functions and comparison
operators belong to the background signature. 
If $I'$ is an interpretation of
signature
$\{\i{Amount}_0, \i{Amount}_1, \i{FillUp}\}$ which agrees with
$I$ on these constants, We say that $I'\models_{bg}\sm[F;\i{Amount}_1]$.
}\medskip

\subsection{Turning ASPMT into SMT for Tight Programs} 

\begin{table*}
\centering 
{\footnotesize
\begin{tabular}{|c|cc|cc|cc|cc|}
\hline 
Instance  & \multicolumn{2}{c|}{ {\sc clingo} v3.0.5 Execution}
& \multicolumn{2}{c|}{iSAT v1.0 Execution} & \multicolumn{2}{c|}{{\sc
    z3 v4.3.0} Execution}  \\ 
Size &  Run Time (Grounding + Solving) & Atoms 
& Run Time (Last step) & Variables & Run time & Memory \\ \hline\hline
10 & 0s (0s + 0s) & 210 & 0s(0s) & 86 & .05s & 2.07 \\ \hline
50 & .02s (.02s + 0s) & 2970 & .05s(0s) & 406 & .18s & 2.17 \\ \hline
100 & .12s (.12s + 0s) & 10920 & .15s(0s) & 806 & .33s & 2.28 \\ \hline
500 & 8.18s (8.17s + 0.01s) & 254520 & 4.41s(.03s) & 4006 & 1.68 &
3.38s \\ \hline
1000 & 55.17s (55.15s + 0.02s) & 1009020 & 18.57s(.09s) & 8006 & 3.35s
& 4.73  \\ \hline
5000 & Did not terminate in 2 hours &  & 500.17s(.45s) & 40006 & 17.42s & 17.32 \\ \hline
10000 & Did not terminate in 2 hours &  & 2008.97s(.93s) & 80006 & 36.49s & 31.42 \\ \hline
\end{tabular}
}
\caption{Leaking Bucket Experiment Results}
\label{table:amtresult}
\vspace{-4mm}
\end{table*}

We say that a formula $F$ is in {\em Clark normal form} (relative to
the list ${\bf c}$ of intensional constants) if it is a conjunction of
sentences of the form 
\beq
   \forall {\bf x} (G\rar p({\bf x}))
\eeq{cnf-p}
and
\beq
\forall {\bf x}y (G \rar f({\bf x}) \mvis y)
\eeq{cnf-f}
one for each intensional predicate $p$ and each intensional function
$f$, where ${\bf x}$ is a list of distinct object variables, $y$ is
an object variable, and $G$ is an arbitrary formula that has no free
variables other than those in ${\bf x}$ and $y$.


The {\em completion} of a formula $F$ in Clark normal form (relative
to ${\bf c}$) is obtained from $F$ by replacing each conjunctive term
(\ref{cnf-p}) with
\[
  \forall {\bf x} (p({\bf x})\lrar G)
\] 
and each conjunctive term (\ref{cnf-f}) with
\[
  \forall {\bf x}y (f({\bf x})\mvis y\lrar G).
\]

An occurrence of a symbol or a subformula in a formula
$F$ is called {\em strictly positive} in $F$ if that occurrence is not
in the antecedent of any implication in $F$.
The {\em t-dependency graph} of $F$ (relative to ${\bf c}$) is the
directed graph that 
\begin{itemize}
\item  has all members of ${\bf c}$ as its vertices, and
\item  has an edge from $c$ to $d$ if, for some strictly positive
  occurrence of $G\rar H$ in~$F$,
  \begin{itemize}
  \item  $c$ has a strictly positive occurrence in~$H$, and
  \item  $d$ has a strictly positive occurrence in~$G$.
  \end{itemize}
\end{itemize}

We say that $F$ is {\em tight} (on {\bf c}) if the t-dependency graph
of $F$ (relative to {\bf c}) is acyclic.
For example, 
\[
   ((p\rar q)\rar r)\rar p
\]
is tight on $\{p,q,r\}$ because its t-dependency graph has only one
edge, which goes from $p$ to $r$. On the other hand, the formula is
not tight according to~\cite{ferraris11stable} because, according to
the definition of a dependency graph in that paper, there is an 
additional edge that goes from $p$ to itself. 



Theorem~12 from~\cite{bartholomew12stable} extended the theorem on
completion from~\cite{ferraris11stable} to allow intensional
functions, but it was restricted to a class of formulas called ${\bf
  c}$-plain formulas. The following theorem generalizes that theorem 
by removing that restriction and by referring to the weaker notion of
tightness as described above. 

\begin{thm}\label{thm:completion}\optional{thm:completion}
For any sentence~$F$ in Clark normal form that is tight on ${\bf c}$,
an interpretation $I$ that satisfies $\exists xy(x \ne y)$ is a model
of $\sm[F;{\bf c}]$ iff $I$ is a model of the completion of $F$
relative to ${\bf c}$.
\end{thm}

\begin{table*}
\centering 
{\footnotesize
\begin{tabular}{|c|cc|cc|cc|}
\hline 
Instance Size & \multicolumn{2}{c|}{{\sc clingo} v3.0.5 Execution}
& \multicolumn{2}{c|}{iSAT v1.0 Execution} & \multicolumn{2}{c|}{{\sc z3 v4.3.0} Execution}\\ 
 &  Run Time (Grounding + Solving) & Atoms 
& Run Time & Variables & Run time & Memory \\  \hline\hline
5 & .02s (.02s + 0s) & 3174 & .03s & 331 & .03s & 2.79 \\ \hline
10 & .3s (.3s + 0s) & 10161 & .19s & 596 & .09s & 4.91 \\ \hline
20 & 9.46s (4.02s + 5.11s) & 36695 & .79s & 1126 & .2s & 8.65 \\ \hline
30 & 42.56s (22.32s + 20.24s) & 77627 & 2.05s & 1656 & .36s & 12.22 \\ \hline
50 & 923.74s (297.26 + 626.48s) & 207706 & 14.35s & 2716 & 1.09s &
20.35 \\ \hline
100 & out of memory &  & 494.77s & 5366 & 5.52s & 43.86 \\ \hline
\end{tabular}
}
\caption{Gears World Experiment Results}
\label{table:gearsresult}
\end{table*}

\medskip\noindent{\bf Example~\ref{ex:1} continued}\ \ 
{\sl 
Formula~\eqref{default} is strongly equivalent to 
\[
  \i{Amount}_1\mvis x\ar \neg\neg(\i{Amount}_1\mvis x)\land
   \i{Amount}_0\mvis x\!+\!1, 
\]
so that formula $F$ in Example~\ref{ex:1} can be turned into Clark
normal form relative to $\i{Amount}_1$:  
\[
\ba {rcl}
  \i{Amount}_1\mvis x &\!\!\!\ar\!\!\! & (\neg\neg(\i{Amount}_1\mvis x)\land
  \i{Amount}_0\mvis x\!+\!1) \\ 
  &  & \lor\ (x\mvis 10\land\i{FillUp}).
\ea
\]
and the completion turns it into
\[
\ba {rcl}
  \i{Amount}_1\mvis x &\!\!\!\lrar\!\!\! & (\neg\neg(\i{Amount}_1\mvis x)\land
  \i{Amount}_0\mvis x\!+\!1) \\ 
  &  & \lor\ (x\mvis 10\land\i{FillUp}).
\ea
\]
Using equality, the formula can be written without mentioning the
variable $x$ as 
\[
\ba c
  (\i{Amount}_0\mvis\i{Amount}_1 + 1) \lor (\i{Amount}_1\mvis 10\land\i{FillUp})
  \\
  \i{FillUp}\rar\i{Amount}_1=10\ .
\ea
\]
}

In the language of iSAT, this formula can be represented as 
{\small
\begin{verbatim}
  (Amt = Amt'+1) or (Amt'=10 and FillUp);
  FillUp -> Amt'=10;
\end{verbatim}
}
\noindent
and in the language of Z3, it can be represented as 
{\small
\begin{verbatim}
  (assert (or (= Amt0 (+ Amt1 1)) 
    (and (= Amt1 10) FillUp)))
  (assert (=> FillUp0 (= Amt1 10))) . 
\end{verbatim}
}
Alternatively, according to the method in \cite{bartholomew12stable},
formula~$F$ in Example~\ref{ex:1} can be turned into the input
language of {\sc gringo} by eliminating intensional functions in favor
of intensional predicates.

Our first experiment has the bucket initially at capacity $5$ and the
goal is to get the bucket to capacity $10$ at a certain fixed
timepoint. The different instance sizes correspond to the maximum
capacity of the bucket and the certain timepoint (they are both the
same in each case). iSAT finds a model of bounded length $k$, where
$k$ starts from $0$ and increases by $1$ until a model is found. The
run time reported is the total cumulative times for $k=0,1,\dots, m$
where $m$ is the instance size. The last step time is for the run when
$k=m$. For other systems, we fixed the length $k=m$ from the beginning.
The results shown in Table~\ref{table:amtresult}
demonstrate that even for a relatively simple domain, ASP suffers a
grounding bottleneck that is not present when using SMT solvers. We
see that the number of atoms for {\sc clingo} increases quadratically
to the instance size while the number of variables for iSAT increases
linearly. 


\NB{Give an AMT example; write completion; turn it into iSAT; Z3} 

\NB{head c-plain is subsumed by this?} 

Next, consider the Gears World domain in which we have two gears, Gear1
with radius 7 and Gear2 with radius 17. Each gear is connected to a
motor that has integral running speeds which can be incremented by 1
using the corresponding action. The gears can also be moved close
together so that both gears spin at the speed of the higher value
(between $\i{M1Speed}\times\i{Radius1}$ or
$\i{M2Speed}\times\i{Radius2}$). The goal is to have Gear1 spinning
at a multiple of Gear2's radius. That multiple is the instance size
in Table~\ref{table:gearsresult}, so for example, instance size 3 means
at the end,
we want Gear1 spinning at a speed of $51 (=3\times 17)$. This domain
can be expressed in the language of ASPMT with the theory of 
integers. Here is a part of the program that governs the speed of a
motor. The speed does not change unless the increase action
happens, in which case it is incremented by 1 unit.
\[
\ba {rcl}
  \i{M1Speed}_t\mvis x &\!\!\ar\!\!& \i{M1Speed}_{t\!-\!1}\mvis x\!-\!1 \land \i{IncreaseM1}_{t\!-\!1} \\
  \i{M1Speed}_t\mvis x &\!\!\ar\!\!& \neg\neg (\i{M1Speed}_t\mvis x)
     \land\i{M1Speed}_{t\!-\!1}\mvis x 
\ea
\]
($t$ is a step counter, which can be represented by an ASP variable to
be grounded).

The ASPMT description of the Gears World is tight 
and can be turned into the input language of SMT solvers by
completion. 
For example, the completion relative to $\i{M1Speed}_t$ is 
\[
\ba l
  \i{M1Speeed}_t\mvis x \lrar (\i{M1Speed}_{t\!-\!1}\mvis x-1 \land \i{IncreaseM1}_{t\!-\!1}) \\
\hspace{2.8cm} \lor(\i{M1Speed}_{t\!-\!1}\mvis x \land \neg\neg
\i{M1Speed}_t\mvis x)\ .
\ea
\]

The Gears World domain can also be computed by ASP solvers by
eliminating the intensional functions in favor of intensional
predicates as described in~\cite{bartholomew12stable}.
Table~\ref{table:gearsresult} compares the two approaches and reveals
that the SMT solvers iSAT and {\sc z3} were able to perform
comparatively very well as the instance size increased.

The experiments were performed on an Intel Core 2 Duo CPU 3.00 GHz
with 4 GB RAM. 

\NB{The domain cannot be expressed in the second group}

\section{Comparison with Other Approaches to ASP Modulo Theories} 
  \label{sec:comparison} 

\subsection{Clingcon programs as a special case of ASPMT}


A {\sl constraint satisfaction problem} (CSP) is a tuple $(V,D,C)$,
where $V$ is a set of {\em constraint variables} with the respective
{\em domains} $D$, and $C$ is a set of {\em constraints} that specify
legal assignments of values in the domains to the constraint variables.

A {\sl clingcon program $\Pi$} with a constraint satisfaction
problem $(V,D,C)$ is a set of rules of the form
\beq
   a\ar B, N, \i{Cn},
\eeq{clingcon-rule}
where $a$ is a propositional atom or $\bot$, $B$ is a set of
positive propositional literals, $N$ is a set of negative
propositional literals, and $\i{Cn}$ is a set of constraints from $C$,
possibly preceded by $\no$.

Clingcon programs can be viewed as ASPMT instances. Below is a
reformulation of the semantics in terms of ASPMT.
We assume that constraints are expressed by ASPMT
sentences of signature $V\cup\sigma^{bg}$, where $V$ is a set of
object constants identified with constraint variables $V$ in $(V,D,C)$,
whose value sorts are identified with domains in $D$; we assume that
$\sigma^{bg}$ is disjoint from $V$ and contains all values in~$D$ as
object constants, and other symbols to represent
constraints, such as $+$, $\times$, and $\ge$.
In other words, we represent a constraint as a formula
$F(v_1,\dots,v_n)$ over $V\cup\sigma^{bg}$ where $F(x_1,\dots,x_n)$
is a formula of the signature $\sigma^{bg}$ and $F(v_1,\dots,v_n)$ is
obtained from $F(x_1,\dots,x_n)$ by substituting the object constants
$(v_1,\dots,v_n)$ in $V$ for $(x_1,\dots,x_n)$.

For any signature $\sigma$ that consists of object constants and
propositional constants, we identify an interpretation $I$ of $\sigma$
as the tuple $\langle I^f,X\rangle$, where $I^f$ is the restriction of
$I$ on the object constants in $\sigma$, and $X$ is a set of
propositional constants in $\sigma$ that are true under $I$.

Given a clingcon program $\Pi$ with $(V,D,C)$, and an interpretation 
$I=\langle I^f,X\rangle$, we define the {\sl constraint reduct of
  $\Pi$ relative to~$X$ and $I^f$} (denoted by $\Pi^X_{I^f}$) as the
set of rules
$
   a \ar B
$ 
for each rule~\eqref{clingcon-rule} is in $\Pi$ such that
 $I^f\models_{bg} \i{Cn}$, and
$X\models N$.
We say that a set $X$ of propositional atoms is a {\sl constraint
  answer set} of $\Pi$ relative to $I^f$ if $X$ is a minimal model
of~$\Pi^X_{I^f}$.

\medskip\noindent{\bf Example~\ref{ex:1} continued}\ \ 
{\sl 
The rules
\[
\ba l
  \i{Amount}_1\!+\!1 =^\$ \i{Amount}_0 \ar\no\ \i{FillUp}, \\
  \i{Amount}_1 =^\$ 10 \ar \i{FillUp}
\ea
\]
are identified with 
\[
\ba {l}
  \bot \ar \no\ \i{FillUp}, \no (\i{Amount}_1\!+\!1 =^\$ \i{Amount}_0) \\
  \bot \ar \i{FillUp}, \no (\i{Amount}_1 =^\$ 10) 
\ea
\]
under the semantics of {clingcon} programs. Consider $I$ in
Example~\ref{ex:1}, which can be represented as $\langle I^f,
X\rangle$ where $I^f$ maps $\i{Amount}_0$ to $6$, and $\i{Amount}_1$
to $5$, and $X=\emptyset$.
$X$ is the constraint answer set relative to $I^f$ because $X$ is the
minimal model of the constraint reduct relative to $X$ and $I^f$,
which is the empty set. 

}\smallskip

Similar to the way that rules are identified as a special
case of formulas~\cite{ferraris11stable}, we identify a clingcon
program $\Pi$ with the conjunction of implications $B\land N\land
\i{Cn}\rar a$ for all rules~\eqref{clingcon-rule} in $\Pi$.
The following theorem tells us that clingcon programs are a special
case of ASPMT, in which the background theory is
specified by $(V,D,C)$, and intensional constants are limited to
propositional constants only, and do not allow function constants.

\begin{thm}\label{thm:clingcon}\optional{thm:clingcon}
Let $\Pi$ be a clingcon program with CSP $(V,D,C)$, let ${\bf p}$ be
the set of all propositional constants occurring in~$\Pi$, and let 
$I$ be an interpretation $\langle I^f,X\rangle$ of signature 
$V\cup {\bf p}$. Set $X$ is a constraint answer set of~$\Pi$ relative
to~$I^f$ iff $I\models_{bg} \sm[\Pi;{\bf p}]$.
\end{thm}


Note that a clingcon program does not allow an atom that consists of
elements from both $V$ and ${\bf p}$. Thus the truth value of any atom
is determined by either $I^f$ or $X$, but not by involving both of
them. This allows loose coupling of an ASP solver and a constraint solver.
On the other hand, \cite{gebser09constraint} sketches a method to
extend clingcon programs to allow predicate constants of positive
arity, possibly containing constraint variables as arguments.
This however leads to some unintuitive cases under the semantics of
{\sc clingcon} programs, as the following example shows.
\begin{lstlisting}
  $domain(100..199).  % Office numbers 
  myoffice(a).        % a is my office number,
  :- myoffice(b).     % and b is not.
  :- not a $==b.      % Nevertheless, a equals b.
\end{lstlisting}
System {\sc clingcon} does not notice that this set of assumptions is 
inconsistent. This is because symbols {\tt a} and {\tt b} in
ASP atoms and the same symbols in the constraint are not related. 
On the other hand, ASPMT, which allows first-order signatures, does
not have this anomaly; there is no stable model under ASPMT.


\NB{If we assume there is no constraint variable in propositional
  atoms, then it's okay, but as soon as it is allowed, there is a
  problem}

\subsection{Comparison with ASP(LC) Programs by Liu {\sl et al.}}

\cite{liu12answer} considers logic programs with linear
constraints, or {\em ASP(LC)} programs, comprised of rules of the form 
\beq
   a\ar B, N, LC
\eeq{nlplc}
where $a$ is a propositional atom or $\bot$, $B$ is a set of positive
propositional literals,  and $N$ is a set of negative propositional
literals, and $LC$ is a set of {\em theory atoms}---linear constraints
of the form 
$ 
  \displaystyle\sum\limits_{i=1}^n (c_i\times x_i)\ \bowtie\ k
$ 
where $\bowtie\in\{\leq, \geq, =\}$, each $x_i$ is an object
constant whose value sort is integers (or reals), and each $c_i$, $k$ is an
integer (or real).

An ASP(LC) program $\Pi$ can be viewed as an ASPMT formula whose 
background theory $bg$ is the theory of integers or the theory of reals. 
Let $\sigma^p$ denote the set of all propositional atoms occurring
in~$\Pi$ and $\sigma^f$ denote all object constants occurring in $\Pi$
that do not belong to the background signature. Theory atoms are
essentially ASPMT formulas of signature~$\sigma^f\cup\sigma^{bg}$.
We identify ASP(LC) program $\Pi$ with the conjunction of ASPMT formulas
$B\land N\land LC\rar a$
for all rules~\eqref{nlplc} in $\Pi$. 

An {\em LJN-intepretation} is a pair $(X,T)$ where 
$X\subseteq\sigma^p$ and $T$ is a subset of theory atoms occurring in
$\Pi$ such that there is some interpretation $I$ of signature
$\sigma^f$ such that $I\models_{bg} T\cup\overline{T}$, 
where $\overline{T}$ is the set of negations of each theory atom
occurring in $\Pi$ but not in $T$. 
An LJN-interpretation $(X,T)$ satisfies an atom $b$ if
$b\in X$, the negation of an atom $not\ c$ if $c \notin X$, and a
theory atom $t$ if $t \in T$. The notion of satisfaction is extended
to other propositional connectives as usual. 

The {\em LJN-reduct} of a program $\Pi$ with respect to an
LJN-interpretation $(X,T)$, denoted by $\Pi^{(X,T)}$,
consists of rules 
$
  a \ar B
$
for each rule~\eqref{nlplc} such that $(X,T)$ satisfies 
$
N\land LC$.
$(X,T)$ is an {\em LJN-answer set} of $\Pi$ if $(X,T)$ satisfies
$\Pi$, and $X$ is the smallest set of atoms satisfying $\Pi^{(X,T)}$.

The following theorem tells us that there is a one-to-many
relationship between LJN-answer sets and the stable models in the
sense of ASPMT.

\begin{thm}\label{thm:niemelafsm}\optional{thm:niemelafsm}
Let $\Pi$ be an ASP(LC) program, and $\sigma^p$ and $\sigma^f$ are
defined as above.
\begin{itemize}
\item[(a)] If $(X,T)$ is an LJN-answer set of $\Pi$, then
  for any interpretation $\langle I^f, X\rangle$ of signature
  $\sigma^p\cup\sigma^f$ such that $I^f\models_{bg} T\cup\overline{T}$, we
  have $\langle I^f, X\rangle\models_{bg}\sm[\Pi;\sigma^p]$.

\item[(b)] For any interpretation $I=\langle I^f, X\rangle$ of signature
  $\sigma^p\cup\sigma^f$, if  $\langle I^f, X\rangle\models_{bg}
  \sm[\Pi;\sigma^p]$, then an LJN-interpretation 
   $(X, T)$ where $$T=\{t\mid \text{$t$ is a theory
    atom in $\Pi$ such that $I^f\models_{bg} t$}\}$$ is an
  LJN-answer set of $\Pi$.
\end{itemize}
\end{thm}

\begin{example}
Let $F$ be
\[
\ba{ll}
a \ar x\!-\!z\!>\!0. \hspace{1cm} & b\ar x\!-\!y\!\le\!0. \\ 
c \ar b,\  y\!-\!z\!\le\!0. & \ar \no\ a. \\
b \ar c.
\ea
\]
The LJN-interpretation 
$L = \langle \{a\},\{x\!-\!z\!>\!0\}\rangle$ is an answer set of~$F$
since 
$\{(x\!-\!z\!>\!0, \neg(x\!-\!y\!\le\!0), \neg(y\!-\!z\!\le\!0)\}$ 
is satisfiable (e.g. take $x^I\mvis 2, y^I\mvis 1, z^I\mvis 0$) and
the set $\{a\}$ is the minimal model satisfying the reduct 
$F^L = (\top \rightarrow a)\land c \rightarrow b$. 
On the other hand the interpretation $I$ such that 
$x^I\mvis 2, y^I\mvis 1, z^I\mvis 0, a^I\mvis\true, b^I\mvis\false,
c^I\mvis\false$ satisfies $I\models_{bg}\sm[F;abc]$.
\end{example}

As with {clingcon} programs, ASP(LC) programs are more restrictive
than ASPMT. ASP(LC) programs do not allow theory atoms in the head of
a rule, and like clingcon programs, cannot express intensional functions.

\section{Conclusion}

In this paper, we related the two lines of research on functions in
answer set programming that originated from different motivations, 
leading to an expressive KR formalism called ASPMT,
which can be efficiently computed by SMT solvers. 
The relationship between ASPMT and SMT is similar to the relationship
between ASP and SAT. We expect that, in addition to completion, many
results known between ASP and SAT can be carried over to the
relationship between ASPMT and SMT.

\section*{Acknowledgements}

We are grateful to Yunsong Meng for useful discussions related
to this paper. We are also grateful to the anonymous
referees for their useful comments. This work was partially supported
by the National Science Foundation under Grant IIS-0916116 and by the
South Korea IT R\&D program MKE/KIAT 2010-TD-300404-001.

\bibliographystyle{named}


\end{document}